# KS_JU@DPIL-FIRE2016:Detecting Paraphrases in Indian Languages Using Multinomial Logistic Regression Model

Kamal Sarkar
Department of Computer Science and Engineering
Jadavpur University, Kolkata, India
jukamal2001@yahoo.com

## ABSTRACT

In this work, we describe a system that detects paraphrases in Indian Languages as part of our participation in the shared Task on detecting paraphrases in Indian Languages (DPIL) organized by *Forum for Information Retrieval Evaluation* (FIRE) in 2016. Our paraphrase detection method uses a multinomial logistic regression model trained with a variety of features which are basically lexical and semantic level similarities between two sentences in a pair. The performance of the system has been evaluated against the test set released for the FIRE 2016 shared task on DPIL. Our systemachieves the highest *f-measure of 0.95* on task1 in *Punjabi* language.The performance of our system ontask1 in Hindi language is *f-measure of 0.90*. Out of 11 teams participated in the shared task, only four teams participated in all four languages, Hindi, Punjabi, Malayalam and Tamil, but the remaining 7 teams participated in one of the four languages. We also participated in task1 and task2 both for all four Indian Languages. The overall average performance of our system including task1 and task2 overall four languages is F1-score of 0.81 which is the second highest score among the four systems thatparticipated in all four languages.

## 1. INTRODUCTION

The concept of paraphrasing is defined in [1] as follows:
"*The concept of paraphrasing is most generally defined on the basis of the principle ofsemantic equivalence: A **paraphrase** is an alternative surface form in the same languageexpressing the same semantic content as the original form.*" Paraphrases may occur at various levels such as lexical paraphrases (synonyms, hyperonymy etc.) , phrasal paraphrase (phrasal fragments sharing the same semantic content) sentential paraphrases ( for example, *I finished my work, I completed my assignment*)[1].

The task of paraphrasing can be of two types based on its applications: paraphrase generation and paraphrase recognition. In broader context, paraphrase generation has various applications. One of the most common applications of paraphrasingis the automatic generation of query variants for submission to information retrieval systems Culicover(1968)[2]describes an earlier approach to query keyword expansionusing paraphrases. The approach in [3] generates several simple variants for compound nouns present in queriesto enhance a technical information retrieval system. In fact, the information retrieval community has extensively exploredthe task of query expansion by applying paraphrasing techniques to generate similar orrelated queries [4][5][6][7][8].

Ravichandran and Hovy (2002)[9] use semi-supervised learning to generate several paraphrasepatterns for each question type and use them in an open-domain question answering system(QA system). Riezler et al. (2007)[10] expand a query by generating *n*-best paraphrases for the queryand then using any novel words in the paraphrases to expand the original query:

NLP applications such as machine translation and multi-document summarization, system performance are evaluated by comparing the system generated output and the references created by human. Manual creation of references is a laborious task. So, many researchers have suggested to use paraphrase generation techniques for generating variants of references for evaluating summarization and machine translation output[11][12].

Callison-Burch, Koehn, and Osborne (2006) [13] use automatically induced paraphrases toimprove a statistical phrase-based machine translation system. Such a system works bydividing the given sentence into phrases and translating each phrase individually bylooking up its translation in a table and using the translation of one of paraphrases of any source phrase that does not have a translation in the table.

Like paraphrase generation, paraphrase recognition is also an important task which is to assign a quantitativemeasurement to the semantic similarity of two phrases [14] or even two given pieces of text[15][16].In other words, the paraphrase recognition task is to detect or recognizewhich sentences in the two texts are paraphrases of each other [17][18][19][20][21][22][23]. The latter formulation of the task has becomepopular in recent years [24] andparaphrase generation techniquesthat canbenefit immensely fromthis task. In general, paraphrase recognition can be very helpfulfor several NLP applications such as text-to-text generationand information extraction.Plagiarism detection is another important application area which needs the paraphrase Identification technique to detect the sentences which are paraphrases of others.

Detecting redundancy is a very important issuefor a multi-document summarization system because two sentences from different documents may convey the same semanticcontent and to make summary more informative, the redundant sentences should not be selected in the summary. Barzilay and McKeown (2005)[25] exploit the redundancy present in a given setof sentences by fusing into a single coherent sentence the sentence segments which are paraphrases of each other. Sekine (2006)[26] shows how to use paraphrase recognition to cluster togetherextraction patterns to improve the cohesion of the extracted information.

Another recently proposed natural language processing task is that of recognizing **textual entailment**: *A piece of text T is said to entail a hypothesis H if humans reading T will infer that H is most likely true* [27][28][29][30].

One of the important requirements for initiating research in paraphrase detection is creation of annotated corpus. The most commonly used corpora for paraphrase detection is the MSRP corpus[1] which contains 5,801 English sentence pairs from news articles manually labelled with 67% paraphrases and 33% non-paraphrases. The shared task on Semantic Textual Similarity conducted as a part of SemEval-2012[2] was targeted to create benchmark datasets for the similar kind of task, but its main focus was to develop systems that can examine the degree of semantic equivalence between two sentences unlike paraphrase detection which determines yes/no decision for given pair of sentences.

However there are at present no annotated corpora or automated semantic interpretation systems available for Indian languages. So creating benchmark data for paraphrases is necessary. With this motivation, creating annotated corpora for paraphrase detection and utilizing that data in open shared task competitions is a commendable effort which will motivate the research community for further research in Indian languages. On this note, the shared task on detecting Paraphrases in Indian Languages (DPIL)@FIRE 2016 is a good effort towards creating benchmark data for paraphrases in Indian Languages. In this shared task, there were two sub-tasks: task1 is to classify a given pair of sentences in Punjabi language as paraphrases (P) or not paraphrases (NP) and task2 is to identify whether a given pair of sentences are completely equivalent (E) or roughly equivalent (RE) or not equivalent (NE). Four Indian Languages –Hindi, Punjabi, Malayalam and Tamil were considered in this shared task. We describe in the subsequent sections our proposed methodology used to implement our system participated in the shared task and we also present performance comparisons of our system with other systems participated in the competition.

## 2. OUR PROPOSED METHODOLOGY

We view the paraphrase detection problem as classification problem. Given a pair of sentences, the task1 is to classify whether the pair of sentences is a paraphrase (P) or not – paraphrase (NP). When task1 is a two class problem, task2 is a three class problem. The task2 is to classify a given pair of sentences into one of three categories: completely equivalent (E) or roughly equivalent (RE) or not equivalent (NE).

Since the problems are basically a classification problem, we have used a traditional classifier for implementing our system. We have used multinomial logistic regression classifier with ridge estimator for both task1 and task2. Each pair of sentences is considered as a training instance. Features are extracted from the training pairs. We consider a number of features for representing sentence pairs. The features which we have used for implementing our system are described in the subsequent subsections:

## 2.1 Features

We have used various similarity measures as the features.

[1] https://www.microsoft.com/en-us/download/confirmation.aspx?id=52398

[2] https://www.cs.york.ac.uk/semeval-2012/task6/index.html

### 2.1.1 Cosine Similarity

To compute cosine similarity, we represent each sentence in a pair using a bag-of-words model. Then cosine similarity is computed between two vectors where each vector corresponds to a sentence in a pair. Basically we consider the set of distinct words in the pair as the vector of features based on which the cosine similarity between two sentences is computed. The size of the vector is *n* where n is $|S_1 \cup S_2|$, $S_1$ is the set of words in the sentence 1 and $S_2$ is the set of words in sentence 2.. Each sentence in a pair is mapped to vector of length *n*. If the vector for sentence 1 is $<v_1, v_2 \ldots v_n>$ and the vector for sentence 2 is $<u_1, u_2 \ldots u_n>$, where $v_i$ and $u_i$ are the values of i-th word feature in sentence 1 and sentence 2 respectively, the cosine similarity between two vectors is computed as follows:

$$Sim_1(S_1, S_2) = cosine(V, U) = \frac{v_1 u_1 + v_2 u_2 + \cdots v_n u_n}{\sqrt{v_1^2 + v_2^2 + \ldots V_n^2} \sqrt{u_1^2 + u_2^2 + \ldots u_n^2}} \quad (1)$$

Here the vector component $v_i$ in vector V corresponds to value of the i-th word feature which is basically the TF*IDF weight of the corresponding word. Similarly vector U is also constructed for the sentence 2.

### 2.1.2 Word Overlap- Exact Match

We also used the word overlap measure as a feature for paraphrase detection. If two sentences in the pair are S1 and S2, the similarity based on word overlap is computed as follows:

$$Sim_2(S_1, S_2) = \frac{|S_1 \cap S_2|}{|S_1| + |S_2|} \quad (2)$$

Where $|S_1 \cap S_2|$ is the number of words common between two sentences. and |S| is the length of sentence S in terms of words.

### 2.1.3 Stemmed Word Overlap

Since the most Indian languages are highly inflectional, stemming is an essential step while comparing words. Accurate stemmers are also not available for Indian languages. So, we applied a lightweight approach to stemming. In this approach, when we match two words, we find the unmatched portions of two words. If we find that the matched portion of two words is greater than or equal to a threshold T1 and the minimum of unmatched portions of word1 and word2 is less than or equal to a threshold T2, we assume that there exists a match between word1 and word2. Stemmed Word overlap is computed using equation (2) with the only difference in word matching criteria. We set T1 to 3 and T2 to 2. We indicate such similarity between two sentences $S_1$ and $S_2$ as $Sim_3(S_1, S_2)$.

### 2.1.4 N-gram Based Similarity

The similarity measures mentioned above compares sentences based on individual word matching. But bag-of-words model does not take into account the context of occurrences of words. We consider n-gram based sentence similarity as one of the features for paraphrase detection.

We compute n-gram based similarity as follows:

$$Sim_4(S_1, S_2) = \frac{c}{a+b} \quad (3)$$

Where

$c$ = # of $n-grams$ matches between $S_1$ and $S_2$ a=# of $n-grams$ in $S_1$

$b$ = # of n grams in $s_2$

We have only considered bigrams(n=2) for implementing our present system.

### 2.1.5 Semantic Similarity

We have used semantic similarity between two sentences as one of the features for paraphrase detection. To compute semantic similarity between sentences, we calculate whether words in the sentences are semantically similar or not. To determine whether two words are semantically similar or not, we have cosine similarity between word vectors for the words. The vector representations of words learned by word2vec models[31] have been used to carry semantic meanings. Word2vec is a group of related models used to produce word embeddings[32] [33]

Word2vec takes as its input a large corpus of text and produces a high-dimensional space where each unique word in the corpus is assigned a corresponding vector in the space.

Such representation of words into vectors positions the word in the vector space such that words that share common contexts are positioned in close proximity to one another in the space

We have used word2vec model available in Python for computing word vectors for the words. We have used *gensim word2vec* model under Python platform with dimension set to 50, min_countto 5(ignore all words with total frequency lower than this). The training algorithm used for developing word2vec model is *CBOW* (Continuous Bag of words). The other parameters of word2vec model are set to default values. If the cosine similarity between the word vectors for the two words is greater than a threshold value, we consider these two words are semantically similar. We set the threshold value to 0.8. We combine a small amount of additional news data with the training data for each language to create the corpus used for computing word vectors. Size of the corpora used to compute word vectors for the different languagesis as follows:

For Hindi, 1.93 MB(8752 sentences), for Punjabi, 1.5 MB(5848 sentences), for Tamil, 2.20 MB (7847 sentences) and for Malayalam, 2.12 MB (7448 sentences)

We compute semantic similarity between two sentences as follows:

$$Sim_5(S_1, S_2) = \frac{e}{f+g} \quad (4)$$

where

$e$ = # of words semantically matches between $S_1$ and $S_2$

f=# words in $S_1$

$g$ = # of words in $s_2$

## 2.2 Our Used Classifier

We have used multinomial logistic regression as the classifier for paraphrase detection task. We view the paraphrase detection problem as a pattern classification problem where each pair of sentences under consideration of paraphrase checking is mapped to a pattern vector based on the features discussed in section 2.1.

We have chosen multinomial logistic regression classifier from WEKA. This is present in WEKA with the name "logistic". WEKA is machine learning workbench consists of many machine learning algorithms for data mining tasks [34].

We set the "ridge" parameter to 0.4 for all our experiments. The other parameters of the classifiers are set to default values.

## 3. EVALUATION AND RESULTS
## 3.1 Description of Datasets

We have obtained the datasets from the organizers of the shared task on detecting paraphrases in Indian Languages (DPIL) held in conjunction with FIRE 2016 @ ISI – Kolkata. The datasets released for four Indian languages-(1) Hindi, (2) Punjabi, (3) Tamil and (4) Malayalam. For each language, two paraphrase detection tasks were defined: Task1- to classify a given pair of sentences in Punjabi language as paraphrases (P) or not paraphrases (NP) and Task2- to identify whether a given pair of sentences are completely equivalent (E) or roughly equivalent (RE) or not equivalent (NE). The training data set for task1 contains a collection of sentence pairs labelled as P (paraphrase) or NP (not a paraphrase) and the training dataset for task2 contains a collection of sentence pairs labelled as completely equivalent (E) or roughly equivalent (RE) or not equivalent (NE). The description of the datasets is shown in Table 1 and Table 2.

**Table 1: Description of Data sets for Task1**

| Language | Training Data Size | Test Data Size |
|---|---|---|
| Hindi | 2500 | 900 |
| Punjabi | 1700 | 500 |
| Malayalam | 2500 | 900 |
| Tamil | 2500 | 900 |

**Table 2: Description of Data sets for Task2**

| Language | Training Data Size | Test Data Size |
|---|---|---|
| Hindi | 3500 | 1400 |
| Punjabi | 2200 | 750 |
| Malayalam | 3500 | 1400 |
| Tamil | 3500 | 1400 |

## 3.2 Evaluation

For evaluating the system performance, two evaluation metrics - Accuracy and F-measure have been used. Accuracy is defined as follows:

$$\text{Accuracy} = \frac{\text{\# of correctly classified Pairs}}{\text{Total \# of Pairs}} \quad (5)$$

Though the same formula was used to calculate accuracy for both the tasks-Task1 and Task2, the formula used to calculate F-measure for Task1 was not the same for Task2. The F-measure used for evaluating task1 is defined as follows:

F1-Score = F1 measure of Detecting Paraphrases=F1- score over P class only.

F-measure for the task2 is defined as:

F1-Score = Macro F1 Score which is an average of F1 scores of all three classes -P, NP and SP.

## 3.3 Results

For system development, we have used training data [35] released for the shared task. At the first stage of this shared task, participants were given the training data sets for system development. At the second stage, the unlabeled test data sets [35] were supplied and the participants were asked to submit the labeled files to the organizers of the contest within a short period of time. Thereafter they evaluated the system output and announced the results. The official results of the various systems participated in Task1 and Task 2 of the contest are shown in Table 3 and Table 4 respectively. As we can see from the tables, no system performs equally well in both the tasks- task1 and task2 acrossall languages. Some systems have performed the best in some languages on task1 and some other systems have performed the best in some other languages on the same task. This is also true for task2.

**Table 3. Official results obtained for Task 1 @ DPIL 2016**

| Team Name | Language | Task | Accuracy | F1 Measure / Macro F1Measure |
|---|---|---|---|---|
| KS_JU | Hindi | Task1 | 0.90666 | 0.9 |
| KS_JU | Malayalam | Task1 | 0.81 | 0.79 |
| **KS_JU** | **Punjabi** | **Task1** | **0.946** | **0.95** |
| KS_JU | Tamil | Task1 | 0.78888 | 0.75 |
| NLP-NITMZ | Hindi | Task1 | 0.91555 | 0.91 |
| NLP-NITMZ | Malayalam | Task1 | 0.83444 | 0.79 |
| NLP-NITMZ | Punjabi | Task1 | 0.942 | 0.94 |
| NLP-NITMZ | Tamil | Task1 | 0.83333 | 0.79 |
| HIT2016 | Hindi | Task1 | 0.89666 | 0.89 |
| **HIT2016** | **Malayalam** | **Task1** | **0.83777** | **0.81** |
| HIT2016 | Punjabi | Task1 | 0.944 | 0.94 |
| **HIT2016** | **Tamil** | **Task1** | **0.82111** | **0.79** |
| JU-NLP | Hindi | Task1 | 0.8222 | 0.74 |
| JU-NLP | Malayalam | Task1 | 0.59 | 0.16 |
| JU-NLP | Punjabi | Task1 | 0.942 | 0.94 |
| JU-NLP | Tamil | Task1 | 0.57555 | 0.09 |
| BITS-PILANI | Hindi | Task1 | 0.89777 | 0.89 |
| DAVPBI | Punjabi | Task1 | 0.938 | 0.94 |
| CUSAT_TEAM | Malayalam | Task1 | 0.80444 | 0.76 |
| ASE | Hindi | Task1 | 0.35888 | 0.34 |
| **NLP@KEC** | **Tamil** | **Task1** | **0.82333** | **0.79** |
| **Anuj** | **Hindi** | **Task1** | **0.92** | **0.91** |
| CUSAT_NLP | Malayalam | Task1 | 0.76222 | 0.75 |

**Table 4. Official results obtainedfor Task 2 by the various participating teams @ DPIL 2016**

| Team Name | Language | Task | Accuracy | F1 Measure/Macro F1 Measure |
|---|---|---|---|---|
| KS_JU | Hindi | Task2 | 0.85214 | 0.84816 |
| KS_JU | Malayalam | Task2 | 0.66142 | 0.65774 |
| KS_JU | Punjabi | Task2 | 0.896 | 0.896 |
| KS_JU | Tamil | Task2 | 0.67357 | 0.66447 |
| NLP-NITMZ | Hindi | Task2 | 0.78571 | 0.76422 |
| NLP-NITMZ | Malayalam | Task2 | 0.62428 | 0.60677 |
| NLP-NITMZ | Punjabi | Task2 | 0.812 | 0.8086 |
| NLP-NITMZ | Tamil | Task2 | 0.65714 | 0.63067 |
| HIT2016 | Hindi | Task2 | 0.9 | 0.89844 |
| **HIT2016** | **Malayalam** | **Task2** | **0.74857** | **0.74597** |
| **HIT2016** | **Punjabi** | **Task2** | **0.92266** | **0.923** |
| **HIT2016** | **Tamil** | **Task2** | **0.755** | **0.73979** |
| JU-NLP | Hindi | Task2 | 0.68571 | 0.6841 |
| JU-NLP | Malayalam | Task2 | 0.42214 | 0.3078 |
| JU-NLP | Punjabi | Task2 | 0.88666 | 0.88664 |
| JU-NLP | Tamil | Task2 | 0.55071 | 0.4319 |
| BITS-PILANI | Hindi | Task2 | 0.71714 | 0.71226 |
| DAVPBI | Punjabi | Task2 | 0.74666 | 0.7274 |
| CUSAT_TEAM | Malayalam | Task2 | 0.50857 | 0.46576 |
| ASE | Hindi | Task2 | 0.35428 | 0.3535 |
| NLP@KEC | Tamil | Task2 | 0.68571 | 0.66739 |
| **Anuj** | **Hindi** | **Task2** | **0.90142** | **0.90001** |
| CUSAT_NLP | Malayalam | Task2 | 0.52071 | 0.51296 |

As we can see from the tables, only 4 teams out of 11 participated teams submitted their systems for all four languages- Hindi, Punjabi, Malayalam and Tamil and the remaining 7 teams participated in only one of the four languages.

We have shown in the tables in bold font the performance scores highest in a particular task for a particular language. It is also evident from the tables that most systems perform well on Punjabi and Hindi languages, but they show relatively poor performance in Tamil and Malayalam languages. We think that the main reason for achieving the better performances inPunjabi and Hindi language domain is the nature of training and testing data sets supplied for those languages. Most likely, that is why most systems perform almost equally well on the Punjabi and Hindi languages. Another reason for having poor performance on Tamil and Malayalam may be the complex morphology of these languages.

We have computed the relative rank order of the participating teams based on overall average performance on task1 and task2 in all four languages (simple average of F1-scores obtained by a team on task1 and task2 over all four languages). Since only four teams have participated in all four languages, we have only shown rank order of these four teams in Table 5.As we can see from Table 5, our system (Team code*: KS_JU*) obtains second best accuracy among the four systems which participated in all four languages.

**Table 5. Overall average performance of systems including task1 and task2 both over all four languages- Hindi, Punjabi, Malayalam and Tamil**

| Team Name | Overall Average F1-Score |
|---|---|
| HIT2016 | 0.84 |
| KS_JU | 0.81 |
| NLP-NITMZ | 0.78 |
| JU-NLP | 0.53 |

## 4. CONCLUSION

In this work, we implement a paraphrase detection system that can detect paraphrases in four Indian Languages-Hindi, Punjabi, Tamil and Malayalam. We use various lexical and semantic level similarity measures for computing features for paraphrase detection task. We view paraphrase detection problem as a classification problem and use multinomial logistic regression model as a classifier. Our model performs relatively better on task1 than on task2.

Our system has the scope for further improvement in the following ways:

- Word2Vec models requires large corpus for proper representation of word meaning, but for our present implementation, we have used a relatively small size corpus for computing word vectors. Use of large corpus for computing word vectors may improve semantic similarity measure leading to improving system performance.
- Since we have only used multinomial logistic regression model as the classifier, there is also the scope to improve the system performance using other classifiers or combination of classifiers.
- Most Indian languages are highly inflectional. So, use of morphological analyzer/stemmer/lemmatizer may improve the system performance.

## 5. ACKNOWLEDGMENTS

This research work has received support from the project entitled ''Design and Development of a System for Querying, Clustering and Summarization for Bengali'' funded by the Department of Science and Technology, Government of India under the SERB scheme.